\documentclass[conference]{IEEEtran}
\IEEEoverridecommandlockouts
\usepackage{tikz} 
\usepackage{cite}
\usepackage{amsmath,amssymb,amsfonts}
\usepackage{algorithmic}
\usepackage{graphicx}
\usepackage{textcomp}
\usepackage{xcolor}
\def\BibTeX{{\rm B\kern-.05em{\sc i\kern-.025em b}\kern-.08em
    T\kern-.1667em\lower.7ex\hbox{E}\kern-.125emX}}
\begin{document}

\title{Leveraging Lecture Content for Improved Feedback: Explorations with GPT-4 and Retrieval Augmented Generation
}

\author{\IEEEauthorblockN{1\textsuperscript{st} Sven Jacobs}
\IEEEauthorblockA{\textit{Computer Science Education} \\
\textit{University of Siegen}\\
Siegen, Germany \\
sven.jacobs@uni-siegen.de}
\and
\IEEEauthorblockN{2\textsuperscript{nd} Steffen Jaschke}
\IEEEauthorblockA{\textit{Computer Science Education} \\
\textit{University of Siegen}\\
Siegen, Germany \\
steffen.jaschke@uni-siegen.de}
}

\newcommand\copyrighttext{%
  \footnotesize \textcopyright \the\year{} IEEE. Personal use of this material is permitted. Permission from IEEE must be obtained for all other uses, in any current or future media, including reprinting/republishing this material for advertising or promotional purposes, creating new collective works, for resale or redistribution to servers or lists, or reuse of any copyrighted component of this work in other works.
  }
\newcommand\copyrightnotice{%
\begin{tikzpicture}[remember picture,overlay]
\node[anchor=south,yshift=10pt] at (current page.south) {\fbox{\parbox{\dimexpr0.75\textwidth-\fboxsep-\fboxrule\relax}{\copyrighttext}}};
\end{tikzpicture}%
}
\newcommand\submittedtext{%
  \footnotesize This work has been accepted for publication by CSEE\&T 2024: 36th International Conference on Software Engineering Education and Training. Copyright may be transferred without notice, after which this version may no longer be accessible.}

\newcommand\submittednotice{%
\begin{tikzpicture}[remember picture,overlay]
\node[anchor=south,yshift=10pt] at (current page.south) {\fbox{\parbox{\dimexpr0.65\textwidth-\fboxsep-\fboxrule\relax}{\submittedtext}}};
\end{tikzpicture}%
}
\renewcommand\fbox{\fcolorbox{red}{white}}
\setlength{\fboxrule}{2pt} 


\maketitle
\copyrightnotice
\begin{abstract}
This paper presents the use of Retrieval Augmented Generation (RAG) to improve the feedback generated by Large Language Models for programming tasks. For this purpose, corresponding lecture recordings were transcribed and made available to the Large Language Model GPT-4 as external knowledge source together with timestamps as metainformation by using RAG. The purpose of this is to prevent hallucinations and to enforce the use of the technical terms and phrases from the lecture. In an exercise platform developed to solve programming problems for an introductory programming lecture, students can request feedback on their solutions generated by GPT-4. For this task GPT-4 receives the students' code solution, the compiler output, the result of unit tests and the relevant passages from the lecture notes available through the use of RAG as additional context. The feedback generated by GPT-4 should guide students to solve problems independently and link to the lecture content, using the time stamps of the transcript as meta-information. In this way, the corresponding lecture videos can be viewed immediately at the corresponding positions.
For the evaluation, students worked with the tool in a workshop and decided for each feedback whether it should be extended by RAG or not. First results based on a questionnaire and the collected usage data show that the use of RAG can improve feedback generation and is preferred by students in some situations. Due to the slower speed of feedback generation, the benefits are situation dependent. 
\end{abstract}


\section{Introduction}
Individual support in teaching and learning contexts with heterogeneous learning groups is desirable in both school and university educational settings, but usually cannot be fully implemented in reality due to the limited availability of teaching staff.  

The topic of individual support by generative AI such as GPT-4 \cite{openai.2023} is particularly promising in computer science education due to its good programming capabilities \cite{bubeck.2023}. Commercially available applications such as ChatGPT, Bard, GitHub Copilot and others are not explicitly designed for skill development or knowledge acquisition, so they directly solve the given programming tasks instead of guiding the learner to solve the problem. This would only be possible with specific prompts. In addition, the external LLM application, such as ChatGPT, must be provided with the current code or error messages each time. For this reason, we have developed the \emph{Tutor Kai} programming exercise  environment with integrated LLM support.

In order to support the student in exercises in the context of an associated lecture with teaching material with feedback, it seems useful if the feedback also refers to the corresponding content and is verifiably linked. This can be achieved by using Retrieval Augmented Generation (RAG), which can also reduce hallucinations because it is based more on real, verifiable facts \cite{lewis.2020}. The question then becomes how to design a feedback system that references and links to lecture information and how students perceive it.

\section{Related Work}
Even before LLMs were available, there were a variety of feedback systems for programming tasks \cite{keuning.2019}. While checking for correct syntax and semantics can be easily automated with unit tests, for example, more specific feedback requires static code checks, which can be time-consuming to set up for each task \cite{jeuring.2022}.

The new possibilities of large language models are manifesting themselves in computer science education in several areas, such as the generation of teaching materials \cite{sarsa.2022} and the analysis of student work \cite{prather.2023}. The CodeAid system, for example, provides students with various programming aids such as Inline Code Exploration, Question from Code, Help Fix Code, Explain Code, and Help Write Code \cite{kazemitabaar.2024}. There are also solutions for automated feedback using LLMs \cite{kiesler.2023a}\cite{hellas.2023}\cite{jacobs.2024a}\cite{phung.2024}, although these do not include specific lecture information in the generated feedback.

While there are still no publications on knowledge-based feedback in programming education, videos are already widely used as a knowledge base for question answering \cite{madasu.2022} and chatbots \cite{wolfel.2024}. Asthana et al. describe a system that uses lecture videos and transcripts as a knowledge base and extracts metadata from them using large language models, e.g. to generate questions about concepts. Feedback is also to be generated on this basis, but has not yet been evaluated \cite{asthana.2023}.

\begin{figure*}[!tbh]
\centerline{\includegraphics[width=\textwidth]{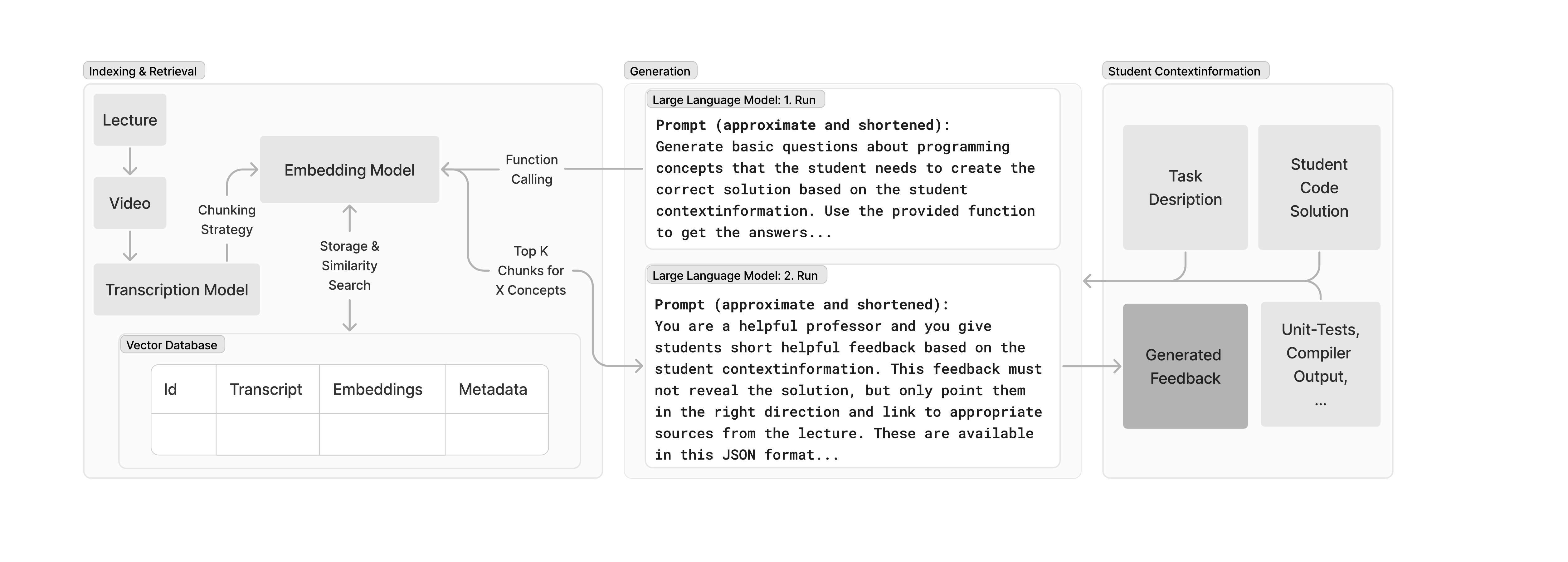}}
\caption{System Design for Enhanced Programming Feedback using Retrieval Augmented Generation}
\label{fig:rag_system}
\end{figure*}

\section{Design and Implementation}

For lectures, the question arises as to which knowledge elements are suitable as a basis. For example, there are texts such as lecture slides, books, and worksheets combined with illustrations. However, the oral explanations of the teacher, which put the content into context, seem to be particularly suitable, since they are available in video form as a lecture recording. By linking the lecture recording to the corresponding timestamp, students are able to perceive the associated visual elements.

\subsection{Indexing}
The lecture recordings were first transcribed into .SRT (SubRip Text) format using the OpenAI speech recognition model Whisper \cite{radford.2023}. This format not only provides the transcribed text, but also includes timestamps for each segment, allowing subsequent linking to specific points in the video. The texts assigned to a segment in the transcript can vary greatly in length, making them unsuitable for further processing. Therefore, a simple chunking strategy has been implemented, which reduces the text assigned to a segment to a uniform size of 512 characters with an overlap of 64 characters to the previous text. The associated timestamp is updated accordingly. A vector representation of the text is stored in a vector database using an embedding model, together with the original text and the start of the associated timestamp, as well as the name of the original video file.  A Postgres database with the pgvector extension and text-embedding-ada-002 from OpenAI is used for Tutor Kai.

\subsection{Retrieval}
In naive RAG, the question is put into a vector representation as a query with the same embedding model for Question Answering, so that similarity scores between the query vector and the vectorized chunks within the indexed corpus can be computed \cite{gao.2024}. The most similar chunks (top K) are then made available to the LLM for answer generation.
However, in order to generate feedback on the solutions to programming tasks, there is no specific question or other text that would be suitable as a query for retrieval in the application. Therefore, a system (Fig. \ref{fig:rag_system}) was implemented that first creates a suitable query for retrieval.

\begin{figure}[b]
\centerline{\includegraphics[width=0.49\textwidth]{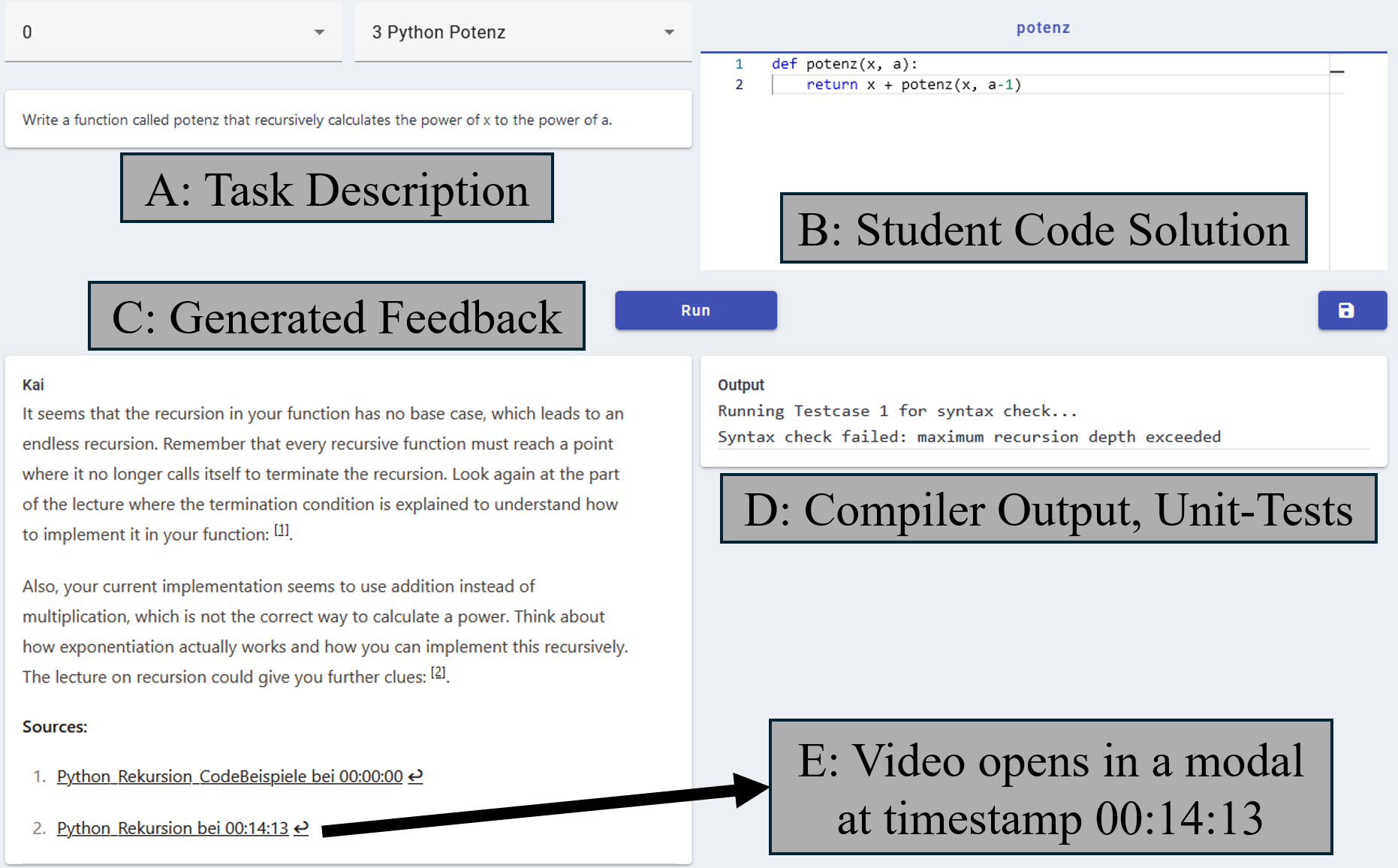}}
\caption{User Interface of Tutor Kai  (translated from German)}
\label{fig:ui}
\end{figure}

\begin{figure*}[!hbt]
\centerline{\includegraphics[width=\textwidth]{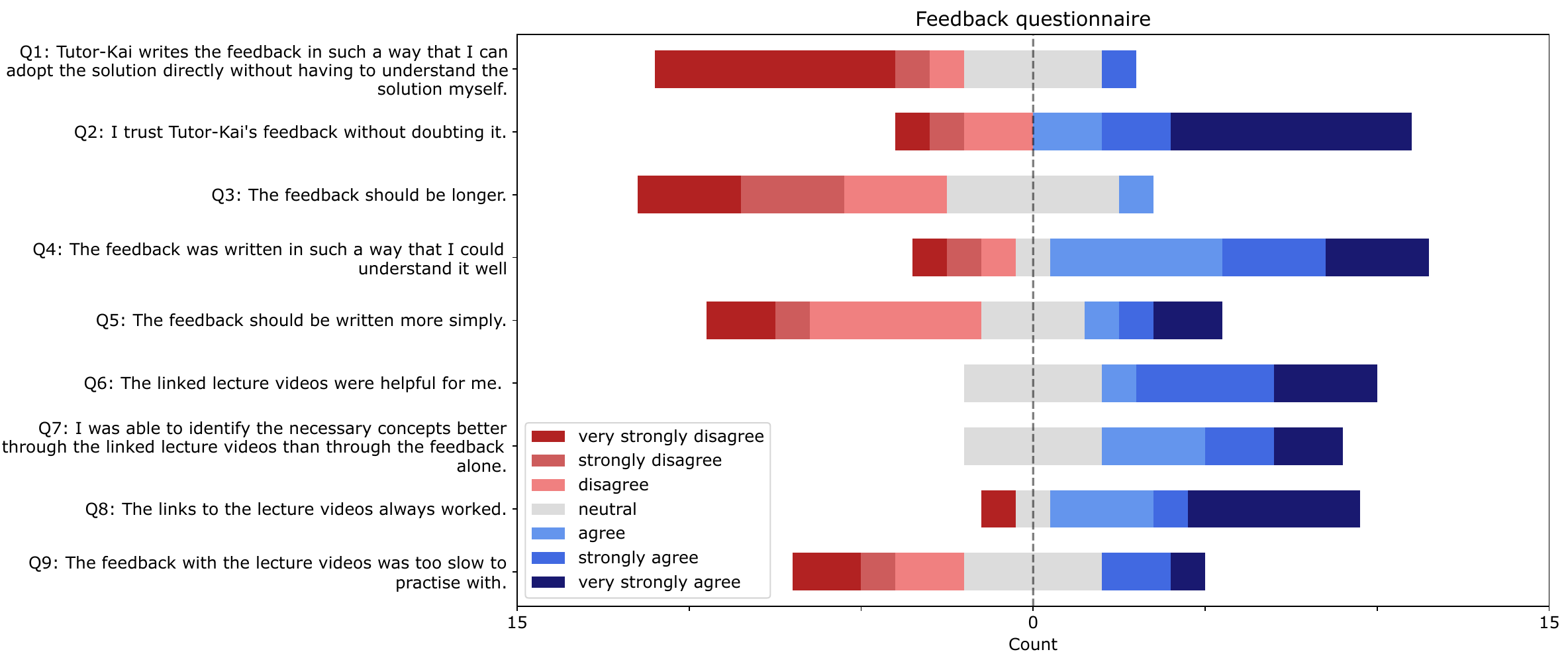}}
\caption{Questionnaire result: Feedback (Questions translated from German)}
\label{fig:feedback}
\end{figure*}

\subsection{Generation}
To implement the logic described, a prompt chain was implemented:
\subsubsection{First Run}
Similarly to the ReAct logic \cite{yao.2023}, the first run identifies X missing concepts for a correct solution based on the available student context information such as task description, student code solution, unit tests and compiler out. For each concept (e.g. recursion), the LLM formulates a simple question (e.g. "How does recursion work in Python?"), which is then used as a query in the described retrieval. 

For Tutor Kai, this is implemented with GPT-4 (1106-preview and temperature = 0) and function calling. A maximum of 2 queries are generated for each feedback, for each of which the top 4 relevant chunks with associated meta information are retrieved. In total, a maximum of 8 relevant chunks with a length of 512 characters each are provided for the second run. In addition, the timestamps and video file names in the metainformation of the retrieved chunks are converted into a Markdown footnote format. In this way, a list of sources is automatically created in the final feedback by a markdown parser.

\begin{figure*}[!hbt]
\centerline{\includegraphics[width=\textwidth]{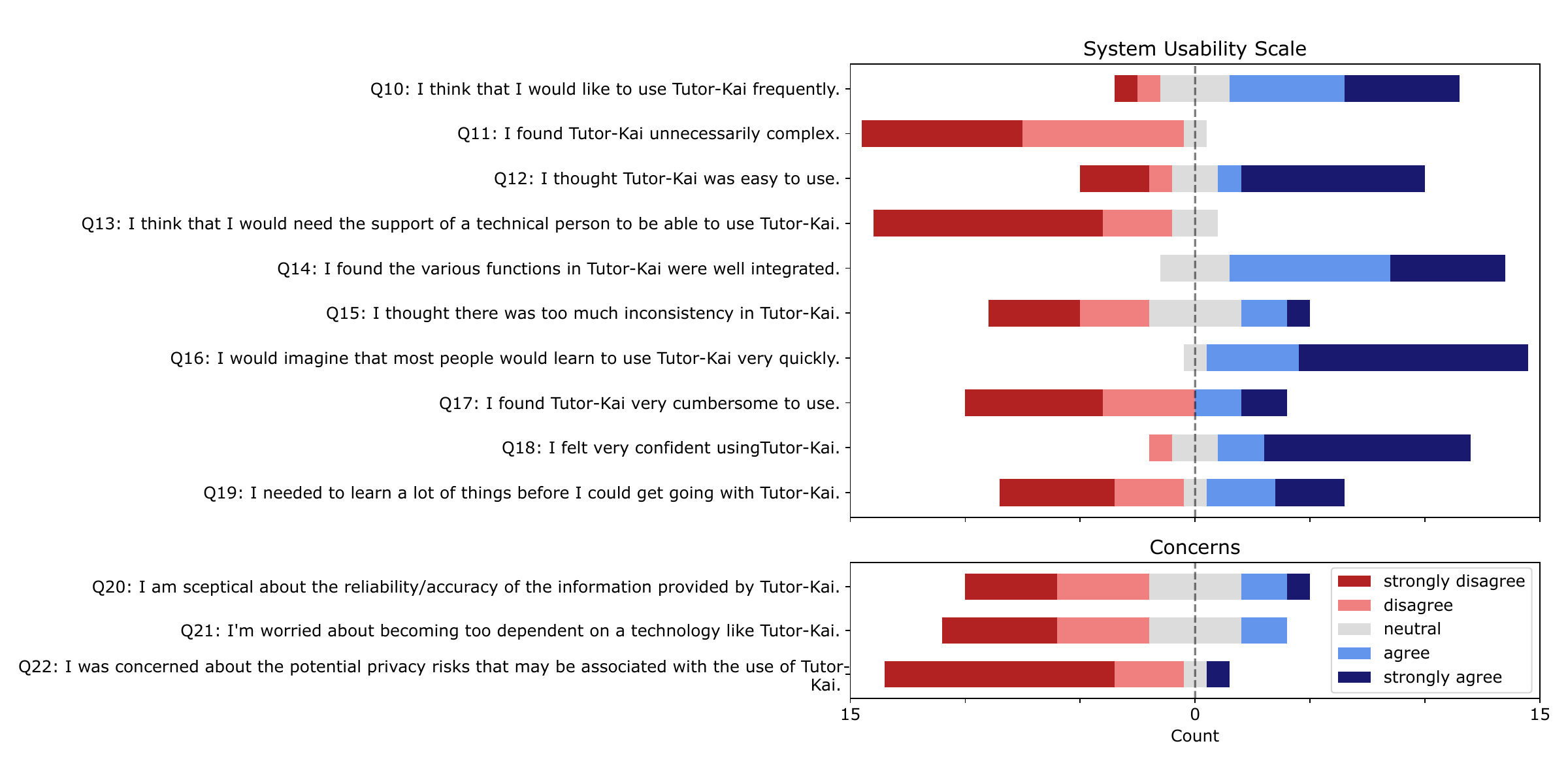}}
\caption{Questionnaire result: System Usability and Concerns (Questions translated from German)}
\label{fig:sus_concerns}
\end{figure*}

\subsubsection{Second Run}
In the second run, the final feedback is generated based on the retrieved chunks and the student context information (Fig. \ref{fig:ui}: A, B, D) using GPT-4 (1106-preview and temperature = 0). The prompt used includes the following elements:
\begin{enumerate}
    \item Role description: The LLM is put in the role of a helpful professor.
    \item Definition of rules: Outputting the solution and formulating the code is prohibited. Feedback should be no more than six sentences in no more than three paragraphs.
    \item Description of the JSON format of the retrieved chunks with associated meta information.
    \item Few Shot Examples on how to cite the retrieved chunks in the final feedback (using the provided markdown footnote link).
    \item Student context information (task description, programming language, student code solution, compiler output, unit test result).
    \item Retrieved lecture chunks in the described JSON format.
\end{enumerate}
Using a markdown parser, linked lecture chunks appear as footnotes in feedback (Fig. \ref{fig:ui}: C). Based on the filename and timestamp, the videos can now be opened in a modal at the linked timestamp (Fig. \ref{fig:ui}: E).

\section{Evaluation}
As part of a voluntary exam preparation workshop, 15 students participated and used Tutor Kai for two to four hours in person and more online over the following three weeks. In addition to the familiar tasks from the previous semester, 10 new tasks were provided. Each time, students could choose between feedback with lecture information and feedback without lecture information. The feedback with lecture information was generated as described. The feedback without lecture information does not use a prompt chain or retrieval, but only the 2nd run (Fig. \ref{fig:rag_system}) without the lecture information (same prompt without elements 3, 4 and 6). 

After the workshop, the opinions of the students (n = 15) about Tutor Kai and the generated feedback were surveyed using a questionnaire. It should be noted that not all students provided responses to every question posed. During and after the workshops, there were 2192 code submissions for which a total of 574 feedbacks were generated. Of these, 478 were feedback without lecture information and 96 were feedback with lecture information generated by the described system.

\subsection{General Evaluation}
An important goal of Tutor Kai is that the students do not receive knowledge of correct result in the feedback, but solve the problem independently, which is ensured by the prompt. This goal is achieved from the students' point of view (Fig. \ref{fig:feedback}: Q1).
Overall, the students are satisfied with both the simplicity and the length of the feedback and were able to understand it well (Fig. \ref{fig:feedback}: Q3, Q4, Q5).

\subsection{Comparing Feedback Types}
In the 96 feedbacks with lecture information, 160 videosegments were linked (average = 1.67). These are spread across 57 different videosegments, with 3 specific segments being linked more than 10 times (maximum = 16).

Lecture information feedback is slower (time to the first streamed token) because the LLM response stream cannot begin until Run 1 (Fig. \ref{fig:rag_system}) is fully completed. The time to the first streamed token depends on the use of the OpenAI API, the length of the task description and the solution of the student's code. Therefore, it cannot be accurately predicted. In our tests, feedback with lecture content took about 18 seconds to stream the first token to the student, while feedback without lecture content took about 1 to 2 seconds.

Students found the additional linking of lecture segments in the feedback helpful and it helped them find the necessary concepts to solve the problem (Fig. \ref{fig:feedback}: Q6, Q7). However, some students found the generated feedback with lecture information too slow (Fig. \ref{fig:feedback}: Q9). This perception is consistent with the additional responses collected in open-ended questions in the questionnaire about why they preferred which type of feedback. In this questionnaire, students mainly referred to the mentioned speed and several times described an approach where feedback without lecture content is generated first, which would be sufficient for "easy" cases, and feedback with lecture information is used for "more difficult" problems.  In the open-ended questions it was mentioned multiple times that the feedback with lecture information helped to remember the lecture.

\subsection{Concerns}
There are concerns that students will become too dependent on LLM-based support systems for programming \cite{prather.2023}. In order to obtain preliminary results, three questions (Q20-Q22) from the TAME-ChatGPT \cite{sallam.2023a} were included in the questionnaire. For the most part, students do not share this concern (Fig. \ref{fig:sus_concerns}: Q21). A possible reason for this is that, unlike ChatGPT and similar applications, Tutor Kai does not provide knowledge of the correct result.

The survey also showed that many students did not question the feedback even though they knew it was generated by generative AI (Fig. \ref{fig:feedback}: Q2 and Fig. \ref{fig:sus_concerns}: Q20). When using such systems, a warning should therefore be displayed at all times.

\subsection{System Usability Scale}
The system used was also evaluated for system usability using the System Usability Scale (Fig. \ref{fig:sus_concerns}) \cite{sauro.2016}. The final usability score was 74.8. Since the user interface consists of only a few clear elements, it should be straightforward to use. It should be noted that students rated their overall experience with Tutor Kai beyond the feedback, which also depends on the tasks to be completed. For example, Q19 (Fig. \ref{fig:sus_concerns}) may have been rated negatively by the students because they feel that they still have a lot to learn in order to complete the tasks successfully.

\section{Limitations and Future Work}
Due to the sample size of n=15, only trends are observable. Further research is required to validate these findings. Knowledge and skill acquisition were not the subject of this evaluation. To investigate the extent to which the linked videos were used, future studies should record how long each video was watched per feedback.

There is great potential for indexing and retrieval improvements for future work. For example, specially created short explanatory videos could be better suited than lecture recordings. Additional lecture content could also be linked. The current chunking strategy, which is simply based on the number of characters, could be improved by semantic chunking strategies or the use of a knowledge graph \cite{abu-rasheed.2024}. When generating the necessary concepts in the first run, this could be done several times and the most frequently selected ones could be used by majority voting. Solutions such as the feedback validation \cite{phung.2024} will significantly improve the results in the future or enable completely new applications. In this context, Nori et al. have shown that more advanced prompting techniques can lead to higher performance gains than the development of an improved foundation model \cite{nori.2023}.

\section{Conclusion}
This work investigates how to design a feedback system for programming tasks that refers to lecture content such as videos and can provide concrete content from this information. A two-run prompt chain is used, in which a query for retrieval augmented generation is generated in a first run with GPT-4. In a second run with GPT-4, the retrieved chunks from the transcribed lecture video are used together with a markdown footnote link containing the timestamp and name of the video. Together with the student's context information, the final feedback is generated, linking the corresponding lecture videos at the respective timestamp.

The system has been evaluated with 15 students. Most of them stated in a questionnaire that they found the feedback with linked lecture information helpful and that it helped them to find relevant concepts more quickly. The feedback without lecture information was preferred by students for quick feedback on what they considered a "simple" problem. The feedback with lecture information takes multiple times longer before the first token is streamed, because the query generation has to be completed first. This is another reason why feedback without lecture information was used about four times more often. The speed to the first character is therefore a trade-off. The students were also satisfied with the length and simplicity of the feedback.

\bibliographystyle{IEEEtran}
\bibliography{2024_csee}

\end{document}